\documentclass{article}
\usepackage{diagbox}
\usepackage{spconf,amsmath,graphicx}
\usepackage{ amssymb }
\usepackage{listings}
\usepackage{xcolor}
\usepackage{amsmath}
\usepackage{subfig}
\usepackage{setspace}
\usepackage{caption}
\usepackage{booktabs}
\usepackage{algpseudocode}
\usepackage{algorithm}  
\usepackage{algorithmicx}
\usepackage{ upgreek }
\usepackage{ gensymb }
\usepackage{ dsfont }
\usepackage{ textcomp }

\usepackage[letterpaper, left=1in, right=1in, bottom=1in, top=1in]{geometry}
\title{Fitness Done Right: a Real-time Intelligent Personal Trainer for Exercise Correction}
\name{Yun Chen, Yiyue Chen, and Zhengzhong Tu}
\address{Department of Electrical and Computer Engeering\\
The University of Texas at Austin}
\begin{document}
%
\maketitle
\begin{abstract}
Keeping fit has been increasingly important for people nowadays. However, people may not get expected exercise results without following professional guidance while hiring personal trainers is expensive. In this paper, an effective real-time system called Fitness Done Right (FDR) is proposed for helping people exercise correctly on their own. The system includes detecting human body parts, recognizing exercise pose and detecting errors for test poses as well as giving correction advice. Generally, two branch multi-stage CNN \cite{cao2016realtime} is used for training data sets in order to learn human body parts and associations. Then, considering two poses, which are plank and squat in our model, we design a detection algorithm, combining Euclidean and angle distances, to determine the pose in the image. Finally, key values for key features of the two poses are computed  correspondingly in the pose error detection part, which helps give correction advice. We conduct our system in real-time situation with error rate down to $1.2\%$, and the screen shots of experimental results are also presented.
\end{abstract}

\section{Introduction}
\label{sec:intro}
\begin{figure*}[!t]
\centering
\includegraphics[width=0.7\textwidth]{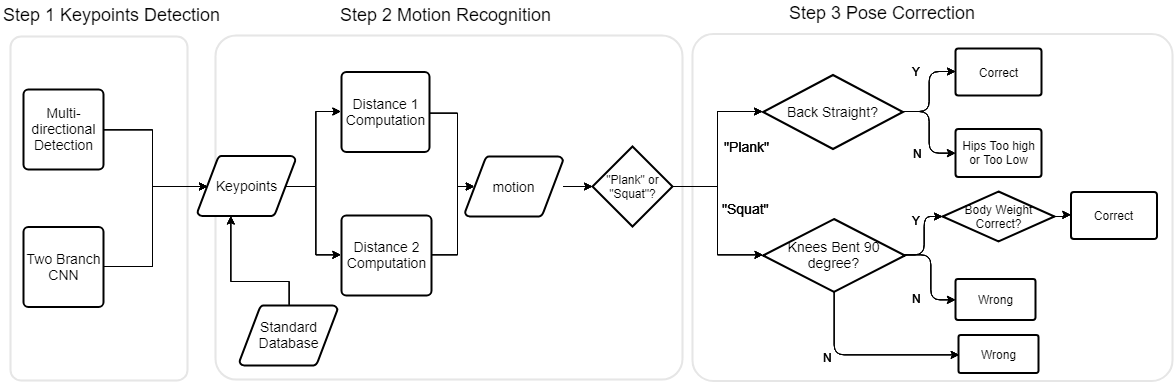}
\caption{Flowchart of FDR.}
\label{Flowchart}
\end{figure*}
Exercise has been increasingly popular in our life, however, since it is expensive and time-consuming to go to gyms and hire personal trainers, people usually work out without precise and professional guidance, which may result in unsatisfying results or muscle hurting. Thus, in order to help people exercise correctly, we propose a practical model for exercise recognition and error correction: Fitness Done Right (FDR). The main goal of FDR is to  detect human body joints, or keypoints, through real-time video stream, then compare them with pictures in standard database for figuring out what the corresponding fitness pose of the test pose, finally, if there are errors in the test pose, correction advice will be offered.\par
As a part of computer vision, lots of pose estimation algorithms are proposed and used in varieties of applications, e.g., \cite{cao2016realtime, simonyan2014very,pishchulin2012articulated, gkioxari2014using,pei2015image,iqbal2016multi,ouyang2014multi,cao2017realtime,ionescu2013human3,liu2019cyclicgen}, most of which use top-down framework to detect human body parts and may suffer from early commitment as errors usually propagate irrecoverably. In addition, none of these models can perfectly fit our scenario for correcting fitness poses. Firstly, a different database should be used for training for our specific goal; Secondly, though it does not really matter if some points are missed in most of the algorithms, it may cause corruption when missing points are essentially keypoints; Finally, photos are taken from different perspectives, which may affects accuracy of angle calculation for both pose recognition and correction parts.\par
In our model,  we followed paper \cite{cao2016realtime} and use two-branch CNN to train and predict keypoints in real-time situations, which is a bottom-up approach and can offer robustness as well as lower complexity compared with top-down approaches. Then, considering two example poses that are plank and squat poses in our model, the severe problem that missing points may result in corruption is solved by rotating frames to several directions until get enough points. Having obtained body part keypoints for each image, we calculate a feature vector for each image and pre-compute standard pose images’ feature vectors to construct the database. The similarity between a test image and standard pose images is derived from a weighted combination of vector’s Euclidean distance and angle distance. From the label of standard image with closest similarity, we can detect the pose for the test image. Knowing the pose from previous steps, pose errors can be found by comparing pose key features between standard fitness pose and the test pose. Least Square approximation is used here for rotating person to ideal degree in 3D dimension and projecting back to 2D dimension.  \par
During the whole process, all of us learned a lot. The whole project used Tensorflow, so we looked into this framework and figured out how it works. We comprehended the two-branch CNN model as well as the algorithms of optimizing the loss functions, particularly AdamOptimizer used in the model. We ran into several problems as mentioned in the second paragraph, but we solved them together and will show them in detail in this report in the following sections, though different problems might come from different parts belonging to one of us. It was really efficient to get together and have discussion about specific problems, and the solutions seemed to come out simply and naturally. \par
The rest of this paper is organized as follows. Section II
introduces a brief framework of FDR, showing whole process of our scheme. Then in Section III, detailed algorithms are provided. The performance of our schemes is evaluated by
experiments in Section IV and Section V concludes our whole project. In Section IV, work layout and our feelings are shown.

\section{Basic Framework}

In this section, a brief framework of FDR including three parts of the whole process is proposed, including keypoints detection, pose recognition and error detection and correction. In addition, some preliminary definitions are provided as well. The whole flow chart is shown in Fig.\ref{Flowchart}.
\subsection{Model training and keypoints detection}
\label{Model training and keypoints detection}
Considering representative joints of human beings, we use 17 keypoints, which are eyes, nose, mouse, ears, shoulders, elbows, wrists, hips, knees and ankles, as defined in \cite{lin2014microsoft}, and limbs are lines connecting these points. The general keypoints and limb graphs can be found in Fig.\ref{transformation}. These keypoints and limbs are trained by using two-branch CNN, which is discussed in detail in Section \ref{Two branch CNN training model} and \ref{Improved body parts and association detection}. 
\subsection{Pose estimation and motion recognition}
\label{Pose estimation and motion recognition}
Given the keypoints from the steps above, we generate a representation vector for each image. Then we use two kinds of vector distance, weighted Euclidean distance and weighted angle distance, to compare the distance between a test image and each image in the standard database. The pose in the test image is derived from that of standard image with smallest distance. More details are discussed in Section \ref{Motion recognition}.
\subsection{Pose error detection and correction}
As mentioned in Section \ref{sec:intro}, people may get unexpected results while training on their own without professional guidance. This part acts like a personal trainer, pointing our the errors of the two poses, e.g., people usually forget to keep straight on their backs when doing plank; it is hard to bend knees correctly and shift their weights to right positions. For these errors, specific detection algorithms are provided as described in Section \ref{Motion_detect}. Then, correction advice will be shown simultaneously on the top left of the screen in order to directly remind the trainer.

\section{detailed algorithms and methods}

In this part, three parts of the system are described in detail: (a) Training model for keypoints detection, which incorporates point auto-augmentation for detecting points that are overlapped or disappeared, and picture rotation for dealing with sparse training data. (b) Similarity calculation between standard motion database and test motion for motion recognition, which uses weighted combination of vector Euclidean distance and angle distance to improve accuracy. (c) Motion feature comparison between standard motions and the test, pointing out right or wrong and giving correction guidance for the wrong motions.
\begin{figure*}[!t]
\centering
\includegraphics[width=0.9\textwidth]{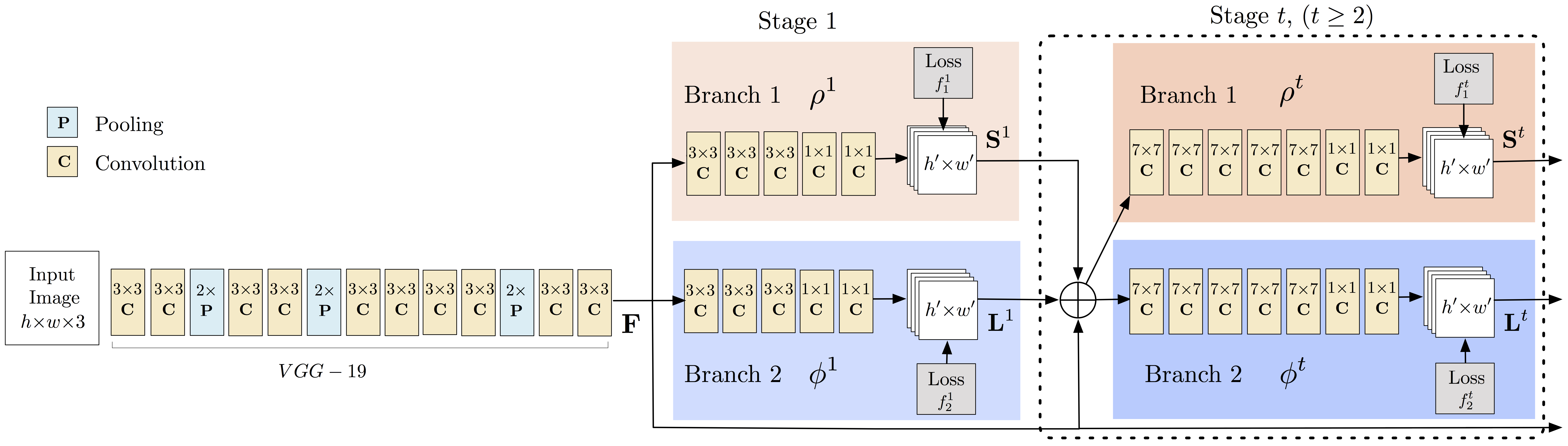}
\caption{Architecture of two-brach multi-stage CNN in \cite{cao2016realtime}.}
\label{Flowchart}
\end{figure*}
\subsection{Two branch CNN model}
\label{Two branch CNN training model}
Following algorithms mentioned in \cite{cao2016realtime}, suppose that input an $w\times h$ image $\rm I$, confidence maps of human body parts is defined as set $S=\{S_1,..., S_j, j\in J\}$, $S_j\in \mathds{R}^{w\times h}$, where $J$ is the set of nodes representing characteristic human body joints; Human joint associations is defined as set $L=\{L_1,...,L_c, c\in C\}, L_c \in \mathds{R}^{w\times h\times 2}$, where $C$ is the set of human joint associations (like limbs), and at each pixel, the 2D vector includes limb position and the orientation of the limb. $S_j$ can be computed by:
\begin{equation}
S_j(x,y) = \exp (-\frac{||(x,y) - (x_j^*,y_j^*)||^2}{\upsigma}),
\end{equation}
where $(x_j^*,y_j^*)$ is the groundtruth position of human body part $j$, and $\upsigma$ controls the spreading rate.\par
The values of $S$ and $L$ at iteration $t$ are:
\begin{equation}
    S^t = \rho ^t ({\rm I}, S^{t-1},L^{t-1});
\end{equation}
\begin{equation}
    L^{t} = \phi ^t({\rm I}, S^{t-1},L^{t-1}).
\end{equation}
where $\rho^t$ and $\phi^t$ should be the respective outputs of two-branch CNN (fine-tuned VGG-19 \cite{simonyan2014very} is used) at either branch at iteration $t$. And the loss function can be defined as:
\begin{equation}
    l^t_S = \sum \limits_{j}\sum \limits_{p = (x,y)} W(p)\cdot ||S^t_j(p) - S^*_j(p)||^2;
\end{equation}
\begin{equation}
    l^t_L = \sum \limits_{c}\sum \limits_{p = (x,y)} W(p)\cdot ||L^t_j(p) - L^*_j(p)||^2.
\end{equation}
where $S^*_j(p)$ and $L^*_j(p)$ are the groundtruth value of $S^*_j$ and $L^*_j$ at position $p$, and $W(p)$ is a binary mask for avoiding penalizing the true positive predictions during traing. In terms of the optimizer, Adaptive Moment Estimation \cite{kingma2014adam} (Adam) is used, which modulates learning rate dynamically by adopting first and second moments estimates of gradient, and thus, learning rate is restricted by $-\frac{\hat{m_t}}{\hat{n_t}+\delta}$, where $\hat{m_t}$ and $\hat{n_t}$ are first and second moments estimates of gradient respectively, and $\delta$ is a small constant for numerical stability. Finally, the loss functions will converge once $l^t_S<\upepsilon _S$ and $l^t_S<\upepsilon _L$. \par
In \cite{cao2016realtime}, multi-person body detection is realized, while in our situation, usually one person is focused to assist exercise correctly, so multi-person parsing using Part Affinity Fields (PAF) is not used in our model.

\subsection{Improved body parts and association detection}
\label{Improved body parts and association detection}

While the two brand CNN model proposed in \cite{cao2016realtime} works quite well in general human pose estimation tasks, it still encounters some difficulties when directly applied to our applications. \par
There are two major challenges we faced while implementing the application. The first issue is low recognition accuracy for none-upright human actions. This is because the training dataset is quite biased for majority of upright poses while lacking relatively ``strange'' poses such as flat position. We should not blame the database itself since the real world poses also centralize on up-right actions such as standing, running, dancing, etc. Thus, we devise two methods to solve this. One is to include more ``strange'' action images in the database and re-train the model to further fit our purpose. This method is quite time-consuming for collecting new images and would also reduce the generalization of the model for the new special-purposed database deviates from the real world scene. Thus we dis-consider this method. \par

\begin{figure}[t]
    \centering
    \subfloat[Direct recognition result.]{
    \includegraphics[width=0.48\columnwidth]{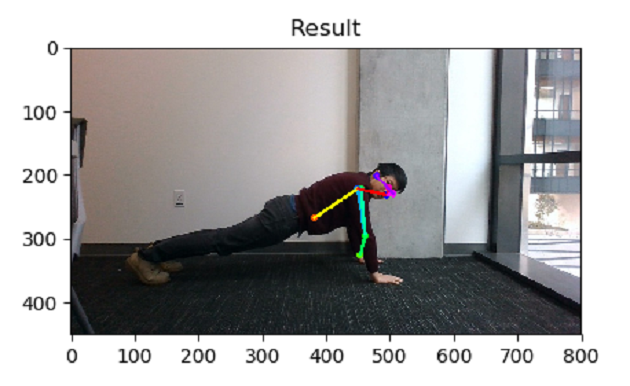}\label{direct_recognition}}\hfill
    \subfloat[Multi-directional recognition result.]{
    \includegraphics[width=0.48\columnwidth]{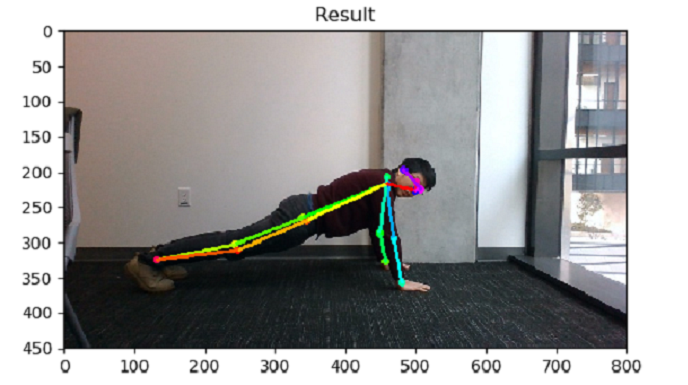}\label{multi_recognition}}\hfill
    \caption{Comparison between direct recognition and multi-directional recognition results on plank pose.  }
    \label{rotation}
\end{figure}

The second method is called multi-directional recognition. The pose image is fed into the inference network with different rotations and the most confident recognition is selected as the final result. This method is simple and effective but doubles the processing time. Considering this trade-off, we choose to rotate by 90\textdegree and it works well for our application. Fig. \ref{rotation} shows the recognition results of multi-directinoal recognition. We can easily see from the figure that the multi-directional recognition greatly improves the recognition accuracy for plank motion. \par

Another challenge is that detected points are usually not sufficient since default model cannot handle overlapping, obscured parts. While it does not necessarily matter for general detection and presentation, it will cause uncertainty in exercise recognition process. So we also added the body part filling module. Based on some heuristic anthropometry, we just copy the corresponding part if some left- or right-part is unrecognizable. For other parts like ``neck'', it is not problematic to average the ``left shoulder'' and ``right shoulder''. Similarly, ``ankle'' can just be extended from the line connecting ``hit'' and ``knee''. These considerations should be as detailed as possible to make sure that all 17 parts are filled properly. Though a little complex, the part filling module works well on improving the matching accuracy in motion recognition (Section \ref{Motion recognition}).\par

\begin{figure*}[!t]
    \centering
    \subfloat[Correct plank pose with straight back and legs.]{
    \includegraphics[width=0.25\textwidth]{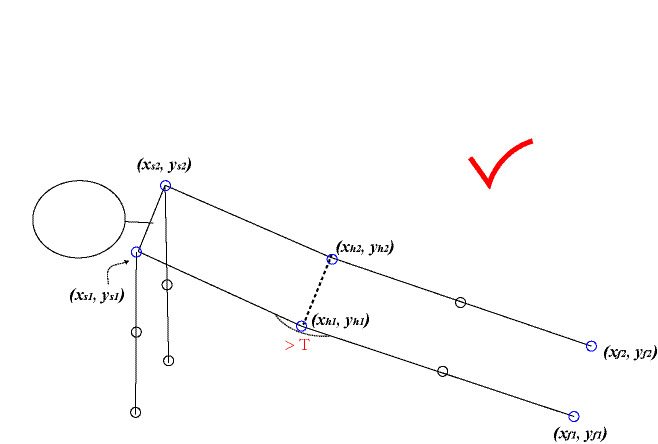}\label{correct_plank}}\hfill
    \subfloat[Wrong plank while hips are too high.]{
    \includegraphics[width=0.25\textwidth]{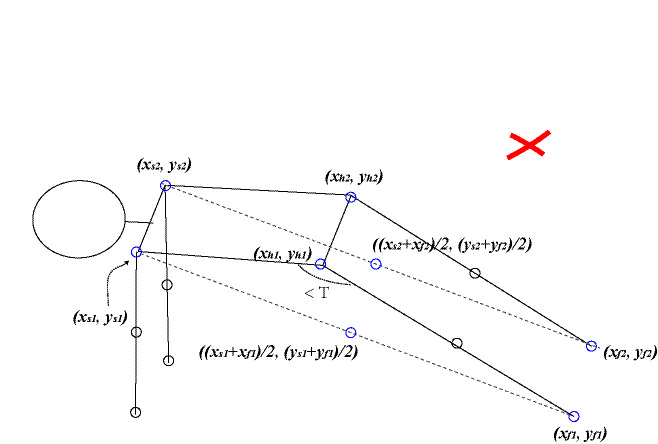}\label{wrong_high}}\hfill
    \subfloat[Wrong plank while hips are too low.]{\includegraphics[width=0.25\textwidth]{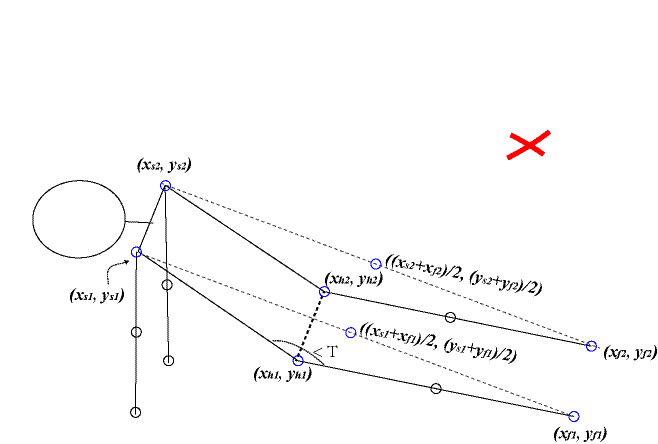}\label{wrong_low}}
    \caption{Typical pose types for plank.}
    \label{wrong_plank}
\end{figure*}
\subsection{Motion recognition}
\label{Motion recognition}
In this step, we manage to identify the motion of the player with the keypoints obtained in the previous steps.

For a given image, we construct its 52-dimension representation vector and 12-dimension angle feature vector. The 52-dimensional vector combines 34 elements for the coordinates of 17 pose keypoints, 17 elements for the confidence score for each keypoint and 1 element for the sum of confidence scores. This is for the calculation of weighted Euclidean distance. For angle feature vectors, we set 12 pairs of body joints, such as left arm-left body and right thigh-hips. Then for each pair of body joints, we compute the cosine distance. The angle feature vector is for the calculation of weighted angle distance.

Before comparing the differences between two images, we then compute their Euclidean distance and angle distance.
Weighted Euclidean distance:
\begin{equation}
d_{E}=\frac{\sum_{k=1}^{17}\beta_{Ak}(|x_{Ak}-x_{Bk}|+|y_{Ak}-y_{Bk}|)}{\sum_{k=1}^{17}\beta_{Ak}}
\end{equation}
where $x_{Ak},x_{Bk},y_{Ak}$ and $y_{Bk}$ represent coordinate of keypoints and $\beta_{Ak}$ represents confidence score of image A.
Weighted angle distance:
\begin{equation}
d_{A}=\frac{\sum_{k=1}^{12}\gamma_{Ak}(|\angle A_{k}-\angle B_{k}|)}{\sum_{k=1}^{12}\gamma_{Ak}}
\end{equation}
where $\angle A_{k} $ and $\angle B_{k} $ represent the k-th element of the angle feature vector of 2 images, and $\gamma_{Ak} $ is taken from the average of confidence scores from 3 keypoints forming the body joint of image A.

The final distance is drawn from a weighted combination of the weighted Euclidean distance and the weighted angle distance. The E-A ratio is the ratio of Euclidean distance to angle distance. In section 4, we test on different values of E-A ratio and set E-A ratio larger than 1. This helps to avoid misdetection of squats when a plank player bending his/her knees.

Considering players’ variance in position of each image, we include angle distance as an improvement for this method. we also preserve the advantages of original Euclidean distance, which gives general position differences of two images, and fine-tune the weights between 2 distance calculation methods to get a more precise detection. 
\subsection{Motion error detection}
\label{Motion_detect}
In this part, basic features of standard exercise are provided, and corresponding angle calculation methods are given in order to decide whether a motion is correct or not.\par
As for plank pose, it is necessary to keep straight on the back, as shown in Fig.\ref{correct_plank}, a plank pose can be defined as correct if:
\begin{equation}
\angle ((x_h-x_s, y_h-y_f),(x_h-x_s, y_h-y_f)) > \mathcal{T}, 
\end{equation}
where $(x_h, y_h), (x_s, y_s), (x_f, y_f)$ are the coordinates of hips, shoulders and feet respectively, and $\mathcal{T}$ is the threshold degree. If the angle is smaller than $\mathcal{T}$, which means the test motion is wrong, two types of typical errors of plank pose are considered, i.e., the hips may be too high or too low, as shown in Fig.\ref{wrong_high} and Fig.\ref{wrong_low}. Thus, correction can be given based on equation as followed:
\begin{equation}
 \left\{ {\begin{array}{*{20}{c}}
{y_h < (y_s+y_f)/2, \quad \quad \text{hips are too high}}\\
{y_h > (y_s+y_f)/2, \quad \quad \text{hips are too low}}
\end{array}.}\right.
\end{equation}\par

As for squat pose, the degree knees are bent needs to be nearly $\pi/2$, and body weights should be shifted to heels, as shown in Fig.\ref{correct_squat}, namely, we define that knees are bent corretly if:
\begin{equation}
\angle ((x_k-x_h,y_k-y_h),(x_k-x_f, y_k-y_h)) \in \frac{\pi}{2}\pm \sigma,
\end{equation}
where $(x_k,y_k), (x_h,y_h), (x_f,y_f)$ are the coordinates of knees, hips and feet respectively, and $\sigma$ is the error range allowed. Moreover, people are required to sit on their back while doing squats, but it is not easy to calculate weight of body accurately, the fraction of horizontal distance between hips and heels and the length of one's thigh is calculated to check whether a person leans too forward, accordingly, body weight position is correct if:
\begin{figure}[!b]
\centering
\includegraphics[width=0.3\columnwidth]{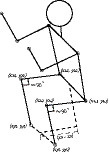}
\caption{Correct squat pose with bending knees to nearly $90\degree$ and shifting body weight to heels.}
\label{correct_squat}
\end{figure}
\begin{equation}
1>\frac{|x_h - x_f|}{|(x_h-x_k,y_h-y_k)|}>\mathcal{F},
\end{equation}
where $\mathcal{F}$ is the threshold deciding the correct body weight position compared with calculated the fraction value.\par
\begin{figure}[!h]
    \centering
    \subfloat[Pose before transformation.]{
    \includegraphics[width=0.48\columnwidth]{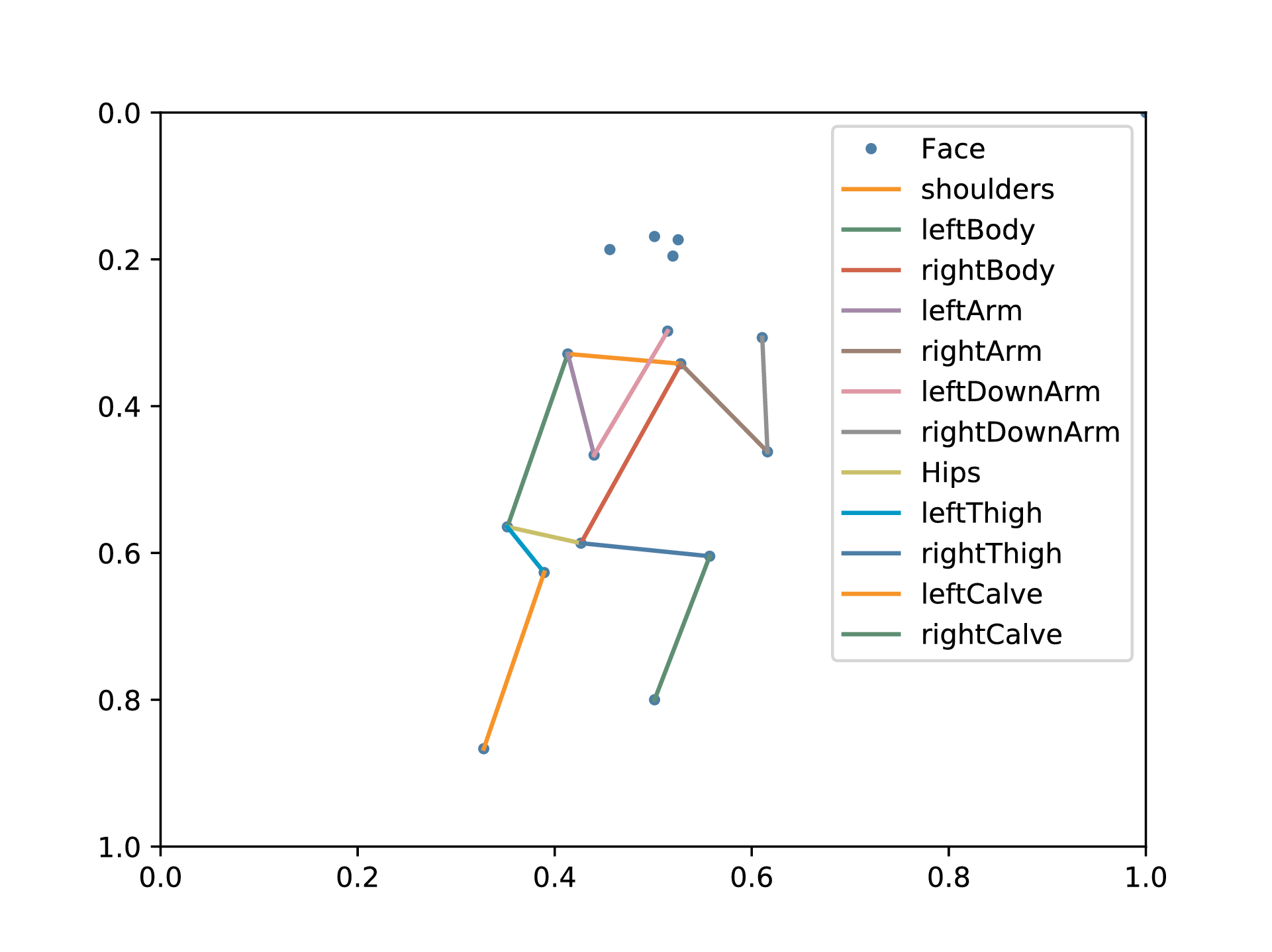}\label{before_transformation}}\hfill
    \subfloat[Refined pose after transformation in 3D dimension.]{
    \includegraphics[width=0.48\columnwidth]{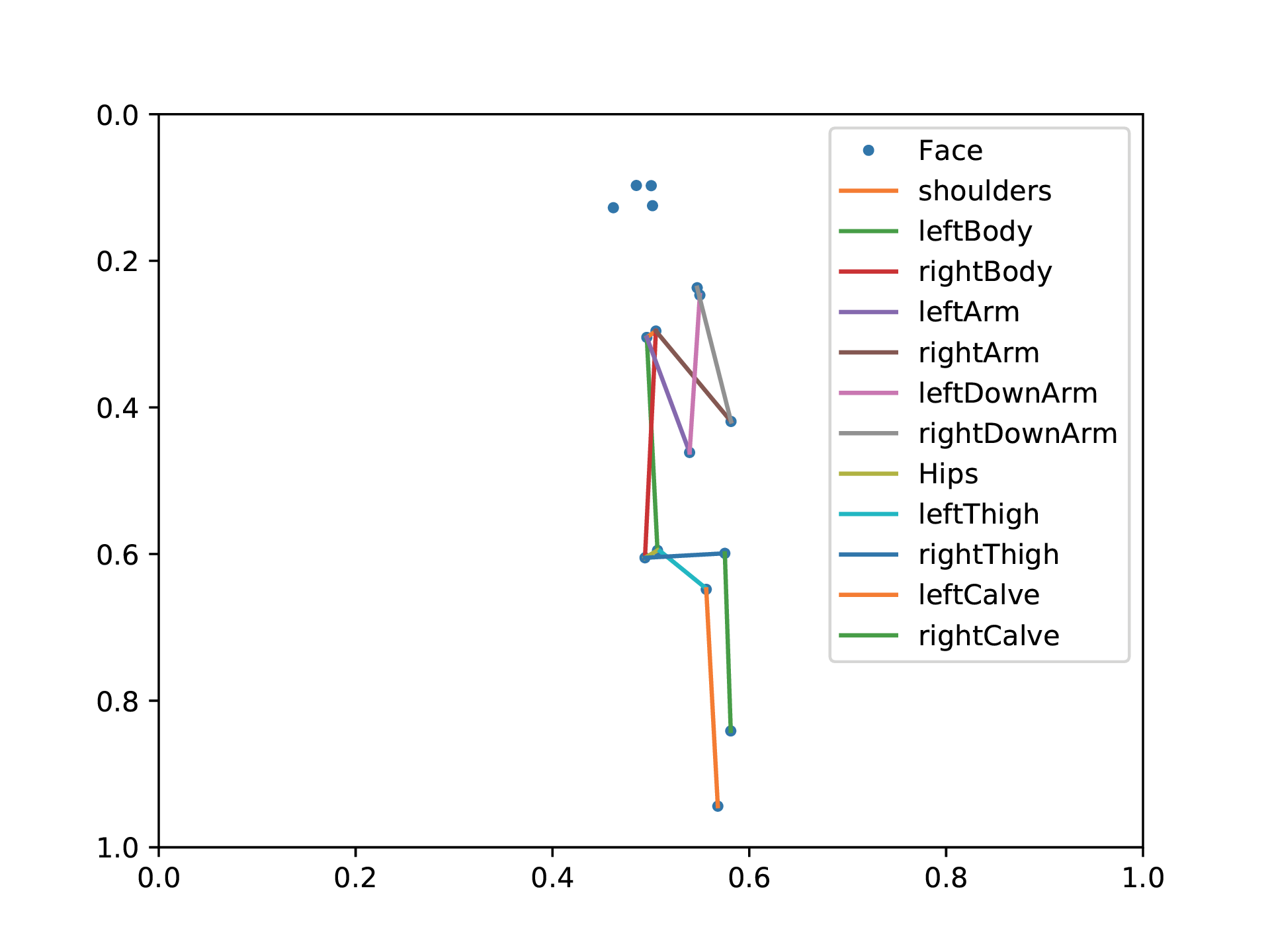}\label{refined_pose}}\hfill
    \caption{Comparison between poses before and after transformation. (a) The angle knees are bent looks obviously less than $90\degree$, so it may be detected as a wrong squat. (b) After transformation, the angle of one knee is nearly $90\degree$, so actually it is correct.  }
    \label{transformation}
\end{figure}
Considering every frame in the video are 2D pictures, which are the projections of objects after rotation, scaling or lens distortion in the 3D dimension possibly, angle calculation of important parts would be affected. In order to get reliable angle from correct perspective, we tried to project 2D coordinates back to corresponding 3D points, then discard the information of third dimension. In our model, Least Square approximation \cite{lai1979strong} is used for getting desired coordinates in both 2D and 3D dimensions. The basic idea is to get projection matrix $\mathds{T}$. Suppose looking from the right or left side, treat human back (keypoints of shoulders and hips) as a 3D rectangle of normalized coordinates $P_{3D} = \{p_{3Di}, i = 1,2,3,4\}$. Define coordinates of human back from a test pose as $P_{2D} = \{p_{2Di}, i = 1,2,3,4\}$, then add pseudo 3D coordinates to $P_{2D}$ to get $\hat{P_{2D}}$. Reorganize $P_{3D}$ and $\hat{P_{2D}}$ as $4\times 3$ matrices $\rm P_{3D}$ and $\hat{\rm P_{2D}}$, thus, the relationship between $\rm P_{3D}$ and $\hat{\rm P_{2D}}$ is:
\begin{equation}
    \rm P_{3D} = \hat{\rm P_{2D}} \cdot \mathds{T}.
\end{equation}
According to least square algorithm, $\mathds{T}$ can be obtained by:
\begin{equation}
\mathds{T} = (\hat{\rm P_{2D}}^{\mathsf{T}}\cdot \hat{\rm P_{2D}})^{-1}\cdot \hat{\rm P_{2D}}^{\mathsf{T}}\cdot \rm P_{3D}.
\end{equation}
The transform process is illustrated intuitively in Fig.\ref{transformation}.

\section{experiment and results}
\begin{figure*}[!t]
    \centering
    \subfloat[Detected correct plank pose.]{
    \includegraphics[width=0.25\textwidth]{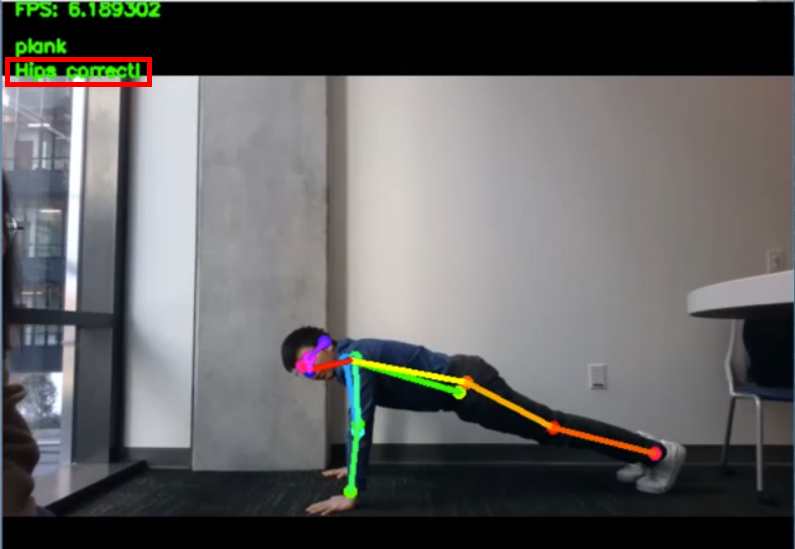}\label{correct_plank}}\hfill
    \subfloat[Wrong plank (hips are too low) with showing error notice.]{
    \includegraphics[width=0.25\textwidth]{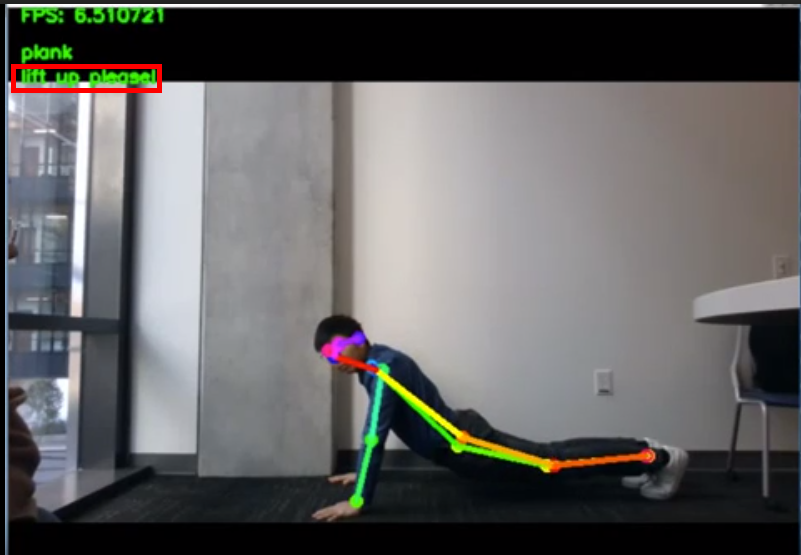}\label{wrong_high}}\hfill
    \subfloat[Wrong plank (hips are too high) with showing error notice.]{\includegraphics[width=0.25\textwidth]{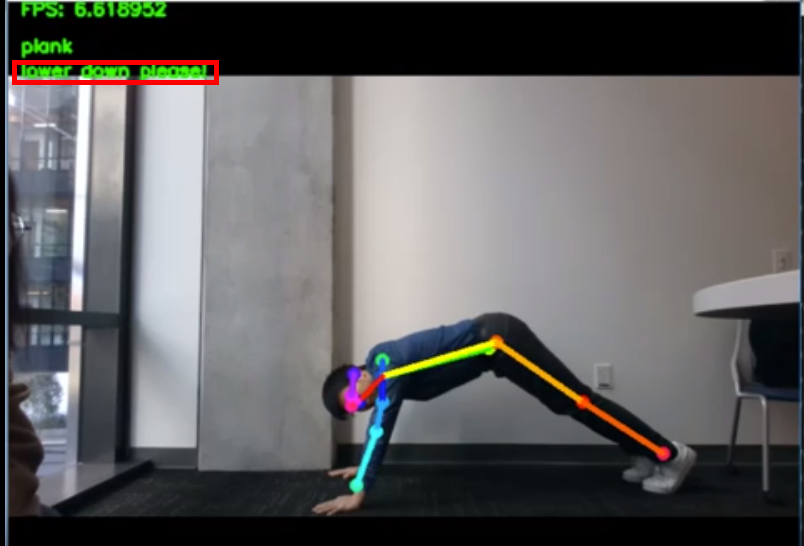}\label{wrong_low}}
    \caption{Comparison among information showed for wrong and correct planks at top left of the video frames.}
    \label{comp_p}
\end{figure*}
\label{sec:typestyle}

Standard motion database construction:
We construct a standard database for detection. The database consisted of 200 images, 90 for planks and 110 for squats. We pre-compute the representation vector and angle feature vector of each image to simplify the following steps.

Motion detection step:
For an image obtained, we compute its representation vector and angle feature vector. Then we calculate distances between this image and each of the database image. The matched motion is the same as the motion of the database image with smallest distance.

\subsection{The selection of E-A ratio}
In our motion detection part, we have combined two distance measuring methods. The Euclidean distance implies the general information on the position of body parts in the image, while the angle distance implies the local angles at body joints, regardless of direction and position of the whole body. Therefore, the value of E-A ratio can affect the detection results, especially when we are detecting planks. In this part, we test on three different values of E-A ratio in plank cases. The error rate is derived from tests over 1000 frames. 

\begin{center}
\begin{tabular}{|c|c|}
\hline E-A ratio&Error Rate\\
\hline 2&1.2\%\\
\hline 1&3.3\%\\
\hline 0.5&5.0\%\\
\hline
\end{tabular}
\end{center}

The error rate decreases as we set larger E-A ratio. This improvement comes from the property of plank: when detecting a plank, we focus more on the direction and relative position of the body, rather than local joint angles. We set E-A ratio equal to 2 in the following experiments.

\subsection{Motion error detection results}
\begin{figure}[!t]
    \centering
    \subfloat[Wrong squat example with error notice shown in red box.]{
    \includegraphics[width=0.48\columnwidth]{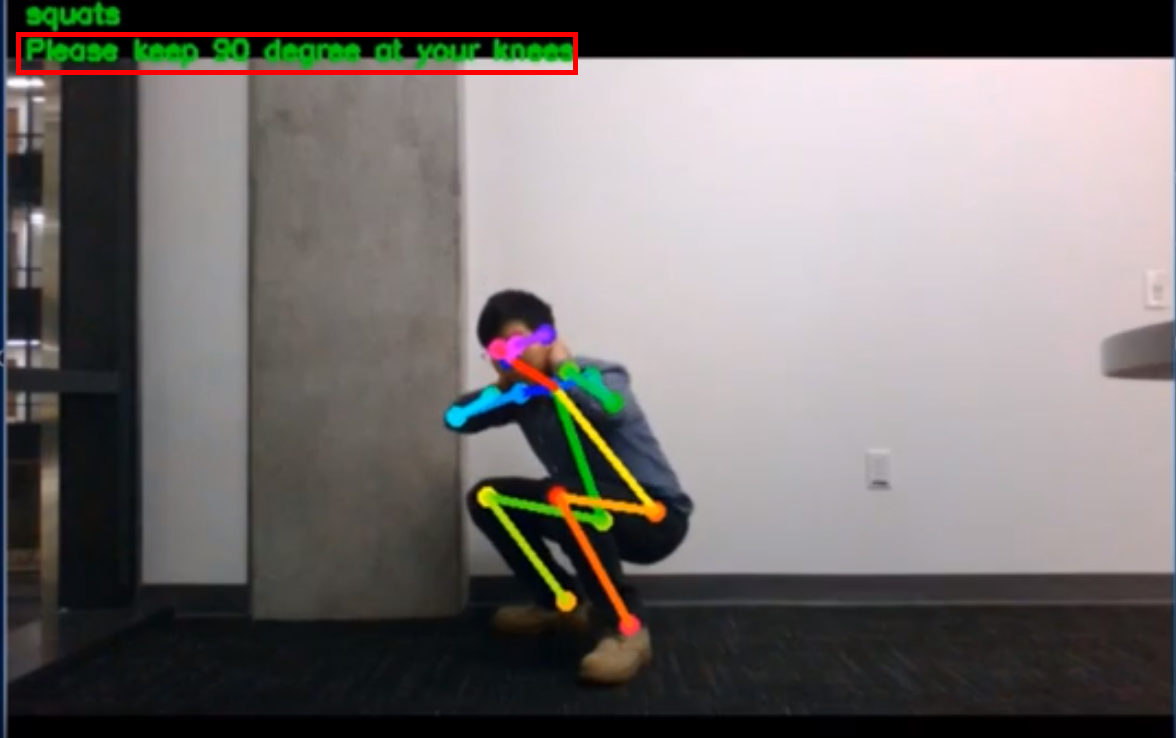}\label{before_transformation}}\hfill
    \subfloat[Correct squat example with correct notice shown in red box.]{
    \includegraphics[width=0.48\columnwidth]{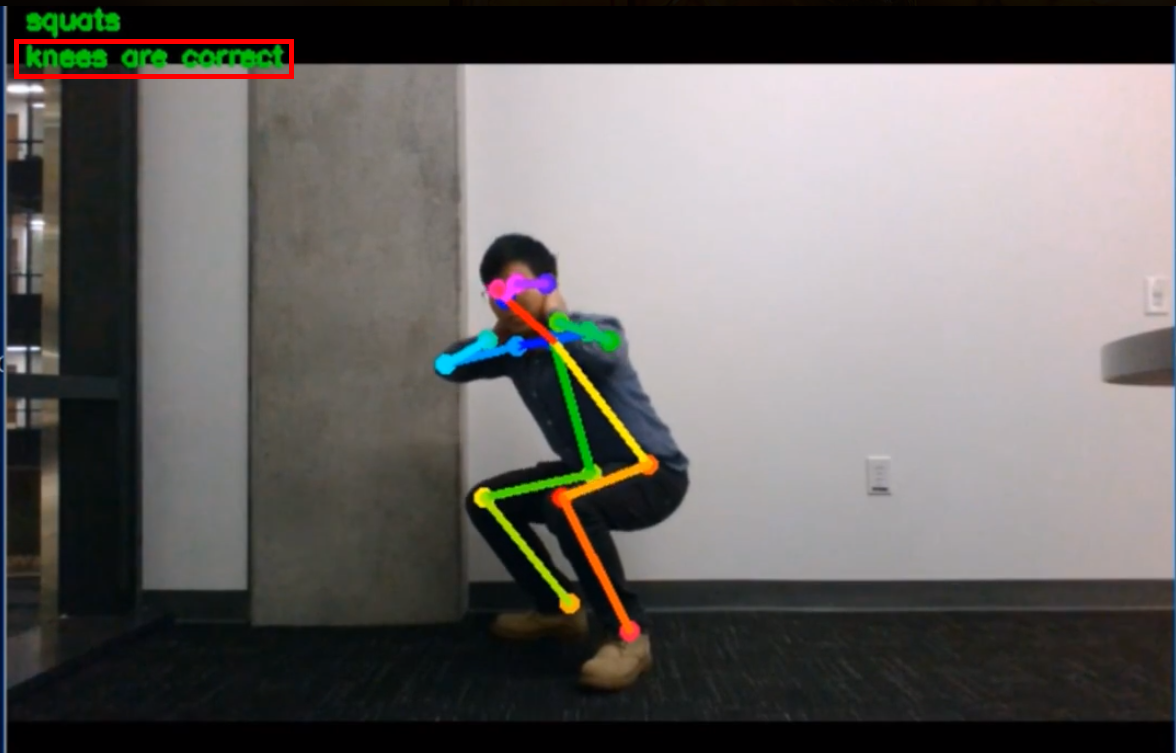}\label{refined_pose}}\hfill
    \caption{Comparison between information showed for wrong and correct squats at top left of the video frames.  }
    \label{comp_s}
\end{figure}

In our model, as described in Section \ref{Motion_detect}, in order to get reasonable results, we set parameters as $\mathcal{T}=165\degree$, $\sigma = 0.05\pi$, $\mathcal{F} = 0.8$. The detection results of squat pose are in Fig.\ref{comp_s}; The results of plank pose are in Fig.\ref{comp_p}. The results show that our model can distinguish correct poses as well as detect errors accurately.

\section{conclusion}
\label{sec:conclusion}
The main goal of this paper is to build a whole system helping people exercise on their own correctly. We divided the system into three parts: human body parts detection, where two-branch CNN combining with multi-directional recognition is used for more accurate detection. Benefiting from robust joints and association detection, in pose recognition part, in order to recognize squat or plank pose, both weighted Euclidean distance and weighted angle distance are adopted and made the recognition error rate down to $1.2\%$. Eventually, considering key features of each pose, Least Square approximation is tried for more accurate angle calculation, and then, pieces of correction advice are given for different situations.\par
Our model can be easily extended to cater for new demands, like detecting more types of poses, we can append standard poses to the database with corresponding pose key features. In short, demands or extensions require only a minor modification of our current system.



\footnotesize
\bibliographystyle{IEEEtran}
\bibliography{IEEEabrv,main}

\end{document}